# A Multiscale Gradient Fusion Method for Edge Detection in Color Images Utilizing the CBM3D Filter

Zhuoyue Wang
Department of Electrical Engineering and Computer Sciences
University of California, Berkeley
Berkeley, USA
zhuoyue_wang@berkeley.edu

Yiyi Tao
Department of Computer Science
Johns Hopkins University
Baltimore, USA
ytao23@jhu.edu

Danqing Ma
School of Informatics, Computing, and Cyber Systems
Northern Arizona University
Flagstaff, USA
madanqing376@gmail.com

Jiajing Chen
Courant Institute
New York University
Sunnyvale, CA
jc12020@nyu.edu

*Abstract*—This paper introduces a novel color edge detection strategy that combines collaborative filtering with multiscale gradient fusion. The approach utilizes the block-matching and 3D (BM3D) filter to enhance sparse representations in the transform domain, effectively reducing noise. To overcome the limitations of single-scale edge detection, particularly the loss of detail, multiscale gradient fusion is employed to improve edge resolution and quality. The process begins by transforming RGB images into the XYZ color space. The colored block-matching and 3D (CBM3D) filter is then applied to reduce noise in the sparse images. Vector gradients of the color images, along with anisotropic Gaussian directional derivatives at two different scales, are calculated and averaged pixel-by-pixel to generate a new edge strength map. To refine the detected edges, image normalization and non-maximum suppression are applied, followed by double threshold selection and a morphological refinement method to extract edge contours. Experimental results on edge detection datasets demonstrate that the proposed method exhibits robust noise resistance and superior edge quality. Comparative evaluations using PR curves, AUC, PSNR, MSE, and FOM metrics show that our approach outperforms existing methods such as Color Sobel, Color Canny, SE, and Color AGDD.

## I. Introduction

Research on grayscale image processing has been ongoing for a long time, and many algorithms in this area are now quite mature. However, with advancements in technology and computer science, color images have become increasingly important, which attracts more interest and research in color image processing in recent years. Common applications of color space include color style transfer [1], color image segmentation [2], color image encryption [3], and color space transformations [17-18], among others.

Edges, as fundamental features in color images, occur where there is an abrupt change in color. Edge detection is essential in tasks like object detection [21], image generation [19], and 3D scene reconstruction [20]. Traditional operators such as Roberts [4], Prewitt [5], and Sobel [6] are effective for grayscale edge detection, but color images offer richer visual information. Wesolkowski et al. [7] enhanced color edge detection by using angle and amplitude measurements, which stabilized gradient calculations. Evans et al. [8] combined mathematical morphology with vector methods for edge detection using vector color differences, though this approach sometimes oversmooths edges, losing detail. Arbelaez et al. [9] used generalized features to encode both local and global contours, yet it can result in blurring and brightness loss. Canny et al. [10] introduced a method with pre-filtering before edge extraction, which remains influential. Deng et al. [11] furthered this approach by merging an improved Sobel operator with wavelet transforms to enhance accuracy, though some fine details may still be lost. Liang et al. [13-14] applied the block-matching and 3D (BM3D) filter to reduce false color information and improve noise robustness while preserving detail.

This paper introduces a novel color edge detection method using multiscale gradient fusion with color BM3D filters. The proposed detector operates in the XYZ color space with a broad color gamut and employs the color BM3D filter for denoising and sparse coding. By combining gradient information with anisotropic directional derivatives at multiple scales, the method constructs a new edge strength map (ESM), which is refined using pixel-average fusion and morphological filling strategies to enhance edge quality. The detector addresses the challenge of large-scale edge elongation while maintaining high resolution and accuracy in small-scale edges, balancing edge quality and noise robustness. Its effectiveness is demonstrated through metrics like PR curve, AUC, PSNR, MSE, and FOM, showing superior performance when compared with Color Sobel [11], Color Canny [12], SE [15], and Color AGDD [16] on various datasets.

## II. Related Work

### A. CIE XYZ color space

The CIE XYZ color space, defined by the International Commission on Illumination (CIE), is an ideal color space based on the three primary colors: red, green, and blue. It was the first color space to be represented mathematically. In this space, the Y component, derived from the green tristimulus value, represents brightness, while the X and Z components correspond to the red and blue stimulus values, respectively.

These three primary components can be combined with positive coefficients to match all perceivable colors.

The brightness information, added to and generated by XYZ, is not contained in the RGB image. An instrument or device is required in RGB color space s, while in the XYZ color space it can be calculated mathematically:

$$\begin{bmatrix} X \\ Y \\ Z \end{bmatrix} = \begin{bmatrix} 0.412453 & 0.357580 & 0.180423 \\ 0.212671 & 0.715160 & 0.072169 \\ 0.019334 & 0.119193 & 0.950227 \end{bmatrix} \begin{bmatrix} R \\ G \\ B \end{bmatrix} \quad (1)$$

As shown in the Equation (1), Z = 0.019334×R + 0.119193×G + 0.950227×B, where the arithmetic sum of the coefficients is 1.088754, which is approximately equal to 1. Assuming the coefficients are normalized so that their sum equals 1, the X, Y, and Z values can then range from 0 to 255.

There is a forward transformation of the XYZ and L*a*b color spaces:

$$\begin{cases} L = 116 f(Y/Y_n) - 16 \\ a = 500 [f(X/X_n) - f(Y/Y_n)] \\ b = 200 [f(Y/Y_n) - f(Z/Z_n)] \end{cases} \quad (2),\text{ where}$$

$$f(t) = \begin{cases} t^{1/3}, \text{if } t > (6/29)^3 \\ \frac{1}{3}\left(\frac{29}{6}\right)^2 t + \frac{16}{116}, \text{if } t \leq (6/29)^3 \end{cases} \quad (3)$$

Among them, the $f(t)$ function has two domains to prevent infinite slope when $t$ equals zero. At $t = t_0$, $f(t)$ is a linear function, that is, $t_0^{1/3} = a t_0 + b$, and its matching slope satisfies $a = 1/(3 t_0^{2/3})$. Figure 1 shows the difference in edge detection between XYZ, RGB, L*a*b and HSV color spaces.

### B. The color block-matching and 3D (CBM3D) filter

An image in XYZ color space with noise variance $\varepsilon_{dB}^2$ can be modeled as $\varphi_{XYZ} = I_{XYZ} + \mu_{XYZ}$, where $I_{XYZ} = [I_X, I_Y, I_Z]$ is actual image or input image and $\mu_{XYZ} = [\mu_X, \mu_Y, \mu_Z]$ is represented as multiplicative or additive noise. Building on the existing BM3D algorithm, the color block-matching and 3D (CBM3D) algorithm can be applied to enhance color image processing. Initially, the XYZ color space image is converted into the luminance-chrominance space, represented by the Y, U, and V channels. In this transformation, the U and V channels predominantly contain low-frequency information, while the Y channel, which exhibits the highest signal-to-noise ratio (SNR), retains the most valuable information, such as texture, edges, and lines. Next, the 2D images of these channels are partitioned into groups, with each block represented as a 3D array. A hard threshold is then applied to the transformed blocks to attenuate noise and shrink the transform spectrum. Finally, an inverse 3D transform is employed to reconstruct the estimates of the grouped blocks, resulting in enhanced image quality.

### III. A NEW COLOR IMAGE EDGE DETECTION METHOD

Assume that the input image in the XYZ color space is $I_{xyz}(x)$ where $x = [u, v]^T$, and the scale is $\sigma$. The expression of the two-dimensional Gaussian kernel function can be defined as:

$$G_\sigma(x) = \frac{1}{2\pi\sigma^2} \exp\left(\frac{-x^T x}{2\sigma^2}\right)$$

$$= \frac{1}{2\pi\sigma^2} \exp\left(-\frac{u^2 + v^2}{2\sigma^2}\right)$$

$$= \frac{1}{\sqrt{2\pi}\sigma} \exp\left(-\frac{u^2}{2\sigma^2}\right) \times \frac{1}{\sqrt{2\pi}\sigma} \exp\left(-\frac{v^2}{2\sigma^2}\right)$$

$$= G_\sigma(u) \times G_\sigma(v) \quad (5)$$

where, $u$ and $v$ are the coordinates of the pixel point, and $T$ represents the transpose of the matrix. For XYZ color images, convolve with a Gaussian function and calculate the gradient magnitude:

$$I_{xyz,\sigma}(x) = I_{xyz}(x) \star G_\sigma(x) = \begin{bmatrix} \iint I_x(x-\delta)G_\sigma(\delta)d\delta \\ \iint I_y(x-\delta)G_\sigma(\delta)d\delta \\ \iint I_z(x-\delta)G_\sigma(\delta)d\delta \end{bmatrix} \quad (6)$$

$$\nabla I_{xyz,\sigma}(x) = \begin{bmatrix} \frac{\partial}{\partial u} I_{xyz,\sigma}(x) \\ \frac{\partial}{\partial v} I_{xyz,\sigma}(x) \end{bmatrix} = \begin{bmatrix} I_x * \nabla G_\sigma(x) \\ I_y * \nabla G_\sigma(x) \\ I_z * \nabla G_\sigma(x) \end{bmatrix} \quad (7)$$

where,

$$\nabla G_\sigma(x) = -\frac{x}{\sigma^2} G_\sigma(x) = \frac{-x}{2\pi\sigma^4} \exp\left(\frac{-x^T x}{2\sigma^2}\right) \quad (8)$$

After obtaining the gradient magnitude of the color space, the composite form formed by each plane and their individual results is computed:

$$\nabla P(x) = \nabla P_{XY}(x) + \nabla P_{YY}(x) + \nabla P_{ZY}(x) \quad (9)$$

$$\begin{cases} \nabla P_{XY}(x) = \nabla P_{XY}(u,v) = \sqrt{[I_x * \nabla G_\sigma(u)]^2 + [I_x * \nabla G_\sigma(v)]^2} \\ \nabla P_{YY}(x) = \nabla P_{YY}(u,v) = \sqrt{[I_y * \nabla G_\sigma(u)]^2 + [I_y * \nabla G_\sigma(v)]^2} \\ \nabla P_{ZY}(x) = \nabla P_{ZY}(u,v) = \sqrt{[I_z * \nabla G_\sigma(u)]^2 + [I_z * \nabla G_\sigma(v)]^2} \end{cases} \quad (10)$$

The result in the composite form is compared with a threshold to obtain the final color gradient magnitude. For formula (5), referring to the anisotropic factor $\rho$, the Gaussian function can be deformed to obtain the anisotropic Gaussian kernel function:

$$G_{\sigma,\rho,\theta}(x) = \frac{1}{2\pi\sigma^2} \exp\left[\frac{-1}{2\sigma^2}\left(x^T R_{-\theta}\right)\begin{bmatrix} \rho^2 & 0 \\ 0 & \rho^{-2} \end{bmatrix}(R_\theta x)\right] \quad (11)$$

$$R_\theta = \begin{bmatrix} \cos\theta & \sin\theta \\ -\sin\theta & \cos\theta \end{bmatrix} \quad (12)$$

where $R_\theta$, a symbol unaffecting the result but only representing the direction of rotation, is the rotation matrix and $\theta$ is the rotation angle. Differentiate the anisotropic Gaussian kernel function to obtain the anisotropic Gaussian directional derivative (AGDD):

$$G'_{\sigma,\rho,\theta}(x) = \frac{(\cos\theta, \sin\theta)x}{\sigma^2 \rho^{-2}} G_{\sigma,\rho,\theta}(x), \theta = (N-1)\pi/n, n = 1,2,3...N \quad (13)$$

N represents the number of directions of AGDD. According to Equations (5) and (13), fuse the color gradient and the anisotropic Gaussian directional derivative gradient of the multi-scale parameter, compute their average gradient value, and obtain a new edge strength map (ESM):

$$\text{ESM} = \max_{n=1,2,3,...N} \left\{ \frac{\left\| |\nabla P(x)| + |I_{xyz} * G'_{\sigma_1,\rho,\theta}(x)| + |I_{xyz} * G'_{\sigma_2,\rho,\theta}(x)| \right\|}{3} \right\} \quad (14)$$

In the XYZ color space, the image is represented as a continuous image. After multiple gradient maps are obtained, they are fused by averaging the gradients pixel by pixel, and the maximum value at each pixel position is used to construct the Edge Strength Map (ESM). To normalize the ESM, the highest pixel value within the ESM is identified, and each pixel value in the ESM is divided by this maximum value, resulting in a normalized ESM.

## IV. EXPERIMENTAL RESULTS AND PERFORMANCE ANALYSIS

To assess the performance and advantages of the proposed color image edge detection algorithm using the CBM3D filter for multiscale gradient fusion, we conducted comparisons with existing algorithms, including Color Sobel [11], Color Canny [12], SE [15], and Color AGDD [16]. These comparisons were performed on both edge detection datasets and non-edge detection datasets. The edge detection datasets, such as BIPED and MDBD, include original RGB images as well as images generated through various image enhancement techniques, such as inversion and translation. In contrast, the non-edge detection datasets consist of images from classic datasets like BSDS500, the PASCAL 2007 challenge dataset, and natural life scenes.

With multiscale factors and edge sharpening operations, the SE algorithm is greatly disturbed by noise, but it provides the advantage of the fastest detection speed among the compared methods.

The Color AGDD algorithm performs well with images containing facial and object features; however, it tends to generate false edges when processing images of low-frequency spatial objects. In contrast, the proposed algorithm effectively detects edges in noise-free conditions and demonstrates superior performance in handling both low-frequency and high-level information images. To evaluate the noise robustness of the algorithms, we conducted experiments by adding zero-mean Gaussian white noise with a noise density of 0.01. The edge detection results of the various algorithms are presented in Figure 1.

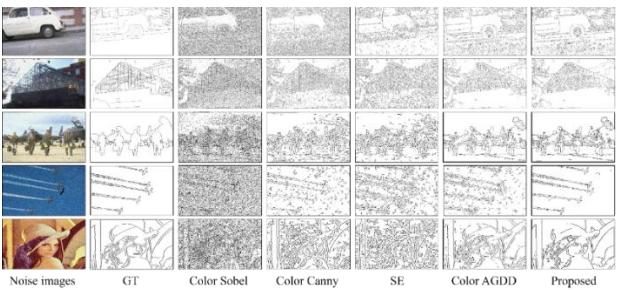

Figure 1. Results of different algorithms with noise density of 0.01

To quantitatively evaluate the performance of the five edge detection algorithms, we used the PR curve, AUC, PSNR, MSE, and FOM indices as metrics. First, the PR curve and AUC were employed to assess the accuracy of each algorithm. In the PR curve, the abscissa (x-axis) represents precision, while the ordinate (y-axis) represents recall. Precision is defined as the proportion of correctly predicted edges among all edges predicted as positive, and recall indicates the proportion of true positive edges that are correctly identified. The formulas for computing precision and recall are as follows:

$$\ell_{\text{precision}} = \frac{\eta_{\text{TP}}}{\eta_{\text{TP}} + \eta_{\text{FP}}}, \quad (17)$$

$$\ell_{\text{recall}} = \frac{\eta_{\text{TP}}}{\eta_{\text{TP}} + \eta_{\text{FN}}}, \quad (18)$$

where $\eta_{\text{TP}}$ is the edge pixel of the true positive, $\eta_{\text{FN}}$ is the edge pixel of the false negative, and $\eta_{\text{FP}}$ indicates that it is not an edge but it is detected. In addition, AUC is defined as the area under PR curve, which can quantitatively reflect the algorithm performance measured based on the PR curve. The value of AUC can be integrated along the horizontal axis by the PR curve:

$$\text{AUC} = \sum_{i=1}^{N_{\text{recall}}} \left\{ \left( \ell_{\text{precision}}(i) + \ell_{\text{precision}}(i+1) \right) \cdot \left( \ell_{\text{recall}}(i+1) - \ell_{\text{recall}}(i) \right) / 2 \right\}, \quad (19)$$

where $N_{\text{recall}}$ represents the number of recall points, and $\ell_{\text{precision}}(i)$ represents the precision value of the $i$-th point. Experiments are carried out on the datasets, and the uniform threshold is 0.001 in the range [0, 1] for positive sequence value. Then, the PR curves of the five algorithms are drawn as shown in Figure 6. Finally, the AUC value calculated according to formula (19) is marked in the figure 2.

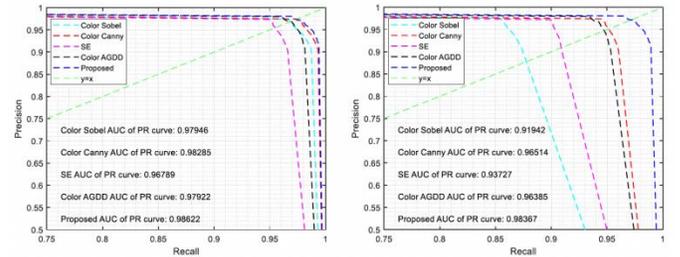

(a) PR curve without noise (b) PR curve with a noise density of 0.01

Figure 2. PR curves and AUC values on the dataset

According to the formula, the values of the five algorithms in the case of PSNR and MSE are obtained and filled in the Table 1.

TABLE 1. PSNR AND MSE VALUES OF THE ALGORITHMS WITH NOISE DENSITY OF 0.01

| PSNR / MSE | Images | | | | |
|---|---|---|---|---|---|
| | Car | House | Airman | Aircraft | Lena |
| Color Sobel | 54.4199/0.2369 | 54.5217/0.2314 | 54.7027/0.2219 | 54.0975/0.2551 | 53.5959/0.2863 |
| Color Canny | 56.4167/0.1496 | 56.6361/0.1422 | 57.5770/0.1145 | 58.2821/0.0973 | 55.7158/0.1758 |
| SE | 58.7171/0.0881 | 56.1659/0.1584 | 57.8254/0.1081 | 56.5962/0.1435 | 55.0433/0.2052 |
| Color AGDD | 58.1061/0.1014 | 57.8259/0.1081 | 58.9112/0.0842 | 58.3481/0.0959 | 57.2148/0.1245 |
| proposed | 59.1469/0.0798 | 58.0401/0.1029 | 59.1629/0.0795 | 60.2004/0.0626 | 57.4664/0.1174 |

TABLE 2. FOM VALUES OF THE ALGORITHMS WITH NOISE DENSITY OF 0.01

| Images | Car | House | Airman | Aircraft | Lena |
|---|---|---|---|---|---|
| $E_{ideal}$ | 899022 | 873012 | 150246 | 155836 | 58613 |
| Color Sobel | 0.7885/712008 | 0.8314/733244 | 0.8048/121460 | 0.9002/142312 | 0.8089/48411 |
| Color Canny | 0.8851/799586 | 0.9353/824910 | 0.9193/138744 | 0.9847/157597 | 0.9479/56721 |
| SE | 0.9472/855264 | 0.9178/809600 | 0.9259/139727 | 0.9484/150229 | 0.9282/55652 |
| Color AGDD | 0.9343/843733 | 0.9254/816180 | 0.9519/143666 | 0.9812/155036 | 0.9818/59257 |
| proposed | 0.9546/861797 | 0.9814/865615 | 0.9558/144224 | 0.9871/159993 | 0.9861/59940 |

Similarly, the Color Sobel and Color Canny algorithms perform poorly in the presence of noise, indicating that neither algorithm is robust enough to handle high noise levels. Under high-intensity noise, the average PSNR of the Color Sobel algorithm is 54.2675, which significantly deviates from the ground truth (GT) image. The SE and Color AGDD algorithms demonstrate a smaller performance gap under noisy conditions and maintain competitive performance. However, the proposed algorithm achieves an average PSNR of 58.8033, which is 4.5358 points higher than the Color Sobel algorithm. Furthermore, the average MSE of the Color Sobel algorithm is 0.2463, which is 15.79% higher than that of the proposed algorithm, as shown in Table 1. In noisy conditions, the quality of all algorithms declines to varying extents, with fluctuations observed in their FOM values. The Color Sobel algorithm performs particularly poorly under high-intensity noise, with an average FOM value of 0.82676 and an average of 351,487 detected edge pixels. While other algorithms show strengths in different datasets, their FOM values are comparable. The proposed algorithm performs best in low-frequency space, achieving a FOM value of 0.9871 and detecting 159,993 edge pixels. Its average FOM value reaches 0.973, which is 14.624% higher than that of the Color Sobel algorithm and 1.808% higher than the next-best Color AGDD algorithm, as detailed in Table 2.

## V. Conclusion

This paper introduces a new approach to color edge detection that combines the CBM3D filter with a multiscale gradient fusion strategy. The method begins by grouping the input image into a sparse matrix, where noise and other high-frequency components are managed using positive and negative 3D transformations alongside collaborative filtering. After converting the image into the XYZ color space, anisotropic directional derivatives of various scales and gradient vectors of the color channels are calculated to create edge maps. These maps are then fused and averaged pixel by pixel to form a new Edge Strength Map (ESM). To refine the ESM, we apply normalization, non-maximum suppression, double thresholding, and morphological filling, which together produce the final edge map.